\documentclass[11pt]{article}

\usepackage[preprint]{acl}

\usepackage{times}
\usepackage{latexsym}

\usepackage[T1]{fontenc}

\usepackage[utf8]{inputenc}

\usepackage{microtype}

\usepackage{inconsolata}

\usepackage{graphicx}
\graphicspath{{../}{./}}

\makeatletter
\def\input@path{{../}{./}}
\makeatother

\usepackage{amsmath}
\usepackage{amssymb}
\usepackage{booktabs}

\usepackage{algorithm}
\usepackage{algorithmic}

\usepackage{listings}
\usepackage{tcolorbox}
\usepackage{xcolor}
\tcbuselibrary{listings, breakable}
\newtcblisting{promptbox}[1][]{
    colback=gray!5,     
    colframe=black!60,    
    breakable,            
    listing only,         
    listing options={
        basicstyle=\ttfamily\small,
        breaklines=true,
        showstringspaces=false,
        mathescape=true
    },
    #1                   
}

\usepackage{enumitem}

\title{From Noisy Traces to Root Causes: Structural Trajectory Analysis and Causal Extraction for Agent Optimization}

\newcommand{\sysname}{{STRACE}}

\author{
 \textbf{Ying Chang\textsuperscript{1 2 \dag}},
 \textbf{Jiahang Xu\textsuperscript{2 *}},
 \textbf{Xuan Feng\textsuperscript{2}},
 \\
 \textbf{Chenyuan Yang\textsuperscript{2}},
 \textbf{Peng Cheng\textsuperscript{2}},
 \textbf{Yuqing Yang\textsuperscript{2}}
\\
 \textsuperscript{1}{University of Chinese Academy of Sciences, Beijing, China},
 \\
 \textsuperscript{2}{Microsoft Research.}
\\
   \textbf{Correspondence:} \href{mailto:jiahangxu@microsoft.com}{jiahangxu@microsoft.com}
}

\begin{document}
\maketitle

\begingroup
\renewcommand{\thefootnote}{\dag}
\footnotetext{Ying Chang did the work during an internship at Microsoft Research.}
\endgroup
\begingroup
\renewcommand{\thefootnote}{*}
\footnotetext{Corresponding author.}
\endgroup

\begin{abstract}

The optimization of long-horizon agents increasingly relies on reflection-based mechanisms, where a large language model (LLM) acts as an optimizer to diagnose agent failures and improve agent policies. However, real execution traces are difficult to use directly for optimization: large trace collections are often redundant and heterogeneous, making optimization inefficient and prone to overfitting to low-value failures; meanwhile, each individual trajectory also contains many irrelevant steps, while naive context reduction methods such as truncation or sliding windows can discard causally important evidence and produce misleading optimization signals.
To resolve this dilemma, we introduce \textbf{STRACE} (\textbf{S}tructural \textbf{Tr}ajectory \textbf{A}nalysis and \textbf{C}ausal \textbf{E}xtraction), a framework that constructs high signal-noise optimization contexts for more precise and effective optimization. At the batch level, STRACE mines failure patterns to filter redundant traces and retain representative failures; within each selected trace, it performs causal localization over a textual dependency graph to remove non-causal steps and identify the true root-cause module for optimization.
%
Empirical results demonstrate that STRACE significantly outperforms standard context-filtering baselines. Notably, on a challenging formal verification task (VeruSAGE-Bench), it successfully optimizes human-expert designed agents, delivering 1.4$\times$ success-rate improvement (42.5\% to 58.5\%). The code 
is available at 
\url{https://github.com/moomight/STRACE}.
\end{abstract}

\section{Introduction}


\begin{figure}[t]
    \includegraphics[width=\columnwidth]{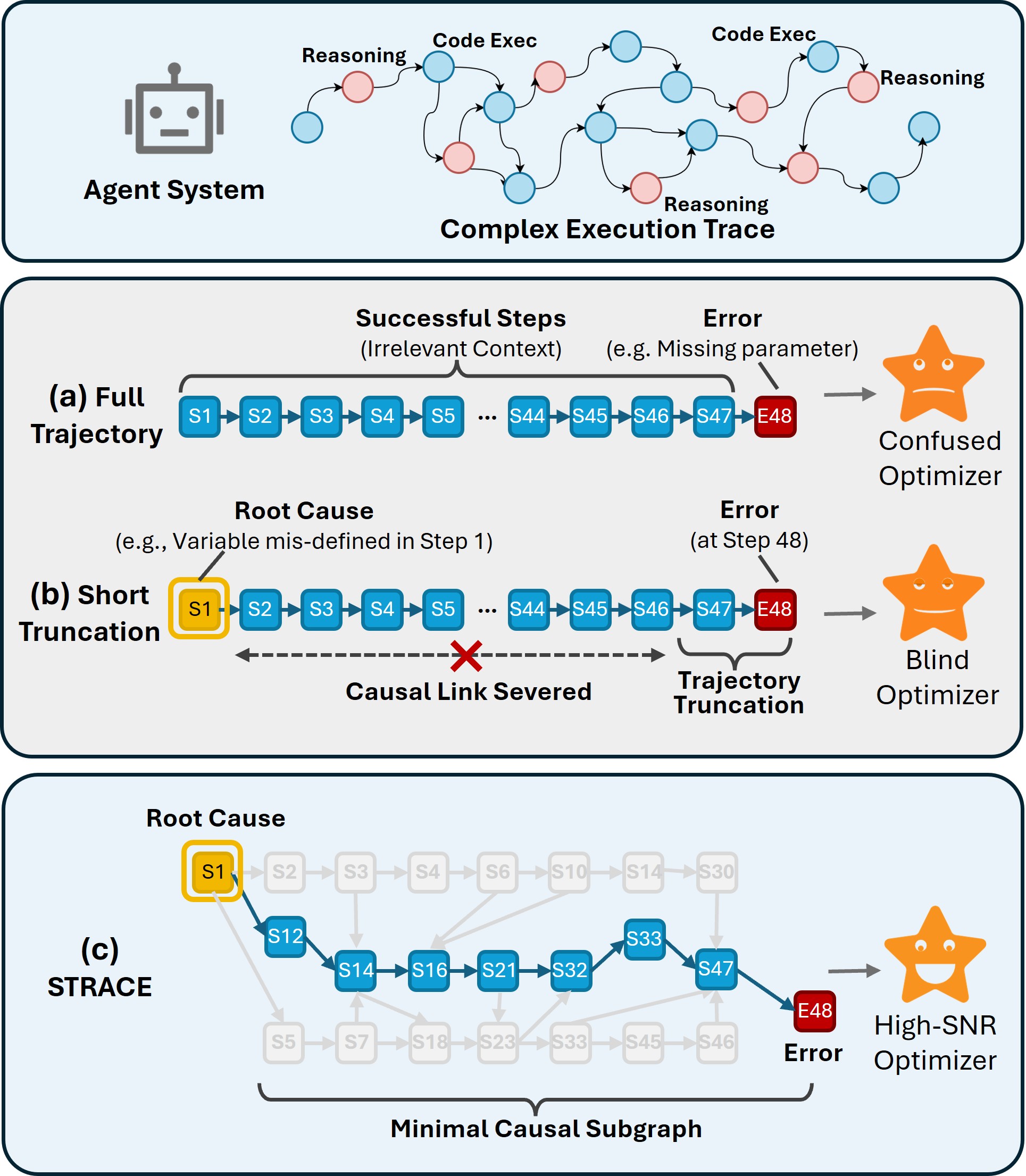}
    \caption{Comparison of context construction strategies. Existing methods struggle with the \textit{context-noise trade-off}: \textbf{(a) Full Trajectory} introduces noise leading to spurious correlations, while \textbf{(b) Short Truncation} fails to capture long-range causal dependencies (e.g., linking Step 1 to Step 48). In contrast, \textbf{(c) STRACE} performs causal context distillation to extract a compact causal slice, providing a high-SNR context for precise optimization.}
\label{fig:teaser}
\vskip -0.2in
\end{figure}

The field of Artificial Intelligence is undergoing a paradigm shift from single-turn interactions to Compound Agent Systems~\cite{compound-ai-blog}. Powered by reasoning-intensive models capable of \textit{Computer Use}~\cite{anthropic2024computer2,anthropic2024computer} and autonomous software engineering~\cite{jimenez2024swebench,chen2021evaluatinglargelanguagemodels,openai2024o1}, these systems are no longer mere chatbots but sophisticated operational units. They orchestrate complex tool usage, manage dynamic memory, and execute multi-step control flows to solve intricate long-horizon tasks~\cite{Wang_2024, xi2023risepotentiallargelanguage, schick2023toolformerlanguagemodelsteach}.
Meanwhile, the artifacts generated by these systems have evolved from simple conversational logs into complex execution trajectories with reasoning chains, code execution environments and state transitions. 
How to effectively use such trajectory information to further enhance agent systems has become a critical question.

A predominant paradigm for optimizing such agent systems is \textbf{\textit{reflexive optimization}}~\cite{pryzant2023automaticpromptoptimizationgradient,shinn2023reflexionlanguageagentsverbal,yang2024largelanguagemodelsoptimizers}, which uses an LLM to diagnose agent-system failures and evolve agent policies based on historical feedback derived from agent trajectories. 
However, when deployed on complex long-horizon tasks, these frameworks struggle to process the sheer volume and complexity of generated data. In realistic settings, batched execution trajectories contain heterogeneous failures with different frequencies, severities, and causes~\cite{cemri2026multi,barke2026agentrx}. Attempting to optimize the agent from every failed trace is computationally intractable and often leads to brittle, overfitted updates. Consequently, a mechanism is needed to aggregate these heterogeneous logs, isolate the most frequent and consequential failure patterns, and distill them into generalized improvements.

Even after representative failures are selected, optimization within an individual trace remains challenging due to the \textit{context-noise trade-off}~\cite{liu2023lostmiddlelanguagemodels}. Existing methods often force a choice between two imperfect extremes (illustrated in Figure~\ref{fig:teaser}). Feeding the \textbf{\textit{full trajectory}} (a) overwhelms the optimizer with irrelevant successful steps and low-SNR context, inviting spurious correlations and hallucinated prompt patches~\cite{yuksekgonul2024textgrad, 10.5555/3618408.3619699}. Conversely, relying on \textbf{\textit{short truncation}} (b) treats local proximity as a signal of causal relevance~\cite{agrawal2025gepareflectivepromptevolution, kang2025aconoptimizingcontextcompression}; it risks discarding distant but causally essential history, severing the link between a symptom and its origin.

This dilemma exposes a fundamental reality of agentic workflows: the node where an error manifests (the downstream symptom) is often not the node where it originates (the upstream root cause)~\cite{zhang2025agentracer}.
For instance, a Code Interpreter crash at Step 50 may be the delayed consequence of an incorrect parameter generated by a Planner at Step 5. Without a mechanism to bridge this temporal gap while filtering noise, optimizers tend to repair symptoms rather than root causes. 

To address these challenges, we introduce \textbf{\sysname{}} (\textbf{S}tructural \textbf{Tr}ajectory \textbf{A}nalysis and \textbf{C}ausal \textbf{E}xtraction), an optimization framework that treats execution logs not as monolithic linear text, but as causal graphs (Figure~\ref{fig:teaser}(c)). \sysname{} systematically resolves these dual challenges through a decoupled pipeline.

First, \textbf{\textit{Structural Modeling}} infers a compact textual dependency graph that records data and control relations among agent components as dependency priors. Building on this topology, \textbf{\textit{Failure Pattern Mining and Trace Filtering}} tackles large-scale trace heterogeneity by summarizing each batch into statistical and structural diagnoses, then filtering redundant or low-value failures using recurrence and severity signals to retain a compact trace set.
Within these representative traces, \textbf{\textit{Causal Localization}} identifies the true root cause as the optimization target and provides the causal context needed for optimization: it traces dependencies backward to discard irrelevant steps, extract a compact causal slice, and locate the failure's logical origin rather than its manifestation node. Finally, during \textbf{\textit{Inductive Policy Optimization}}, \sysname{} uses localized failure episodes to synthesize generalized, natural language heuristics. These preventative guidelines are injected exclusively into the 
instructions
of the identified root-cause modules, enabling continuous, safe, and cost-effective improvement without altering the underlying executable code.

We validate this capability on \textbf{HotpotQA}~\cite{yang2018hotpotqa}, \textbf{WebArena}~\cite{zhou2024webarena}, and \textbf{VeruSAGE-Bench}~\cite{verusage} benchmarks, where \sysname{} significantly outperforms state-of-the-art baselines, notably boosting success rates on the rigorous VeruSAGE benchmark by an absolute \textbf{16.0\%} (42.5\% $\to$ 58.5\%). In summary, our contributions are:
\begin{itemize}
    \item We propose \sysname{}, an end-to-end agent optimization framework that treats execution logs as structured causal evidence rather than monolithic linear text.
    \item We introduce a trace filtering and causal localization mechanism that selects representative failures, extracts compact causal slices, and identifies root-cause modules as optimization targets.
    \item We evaluate \sysname{} on comprehensive benchmarks covering diverse reasoning and long-horizon agent settings, showing consistent gains over full-trajectory, truncation, and prompt-optimization baselines.
\end{itemize}

\section{Related Work}

\subsection{Automated Prompt Optimization}

Automated Prompt Optimization (APO) employs LLMs to iteratively search for superior instructions. Pioneering works formulate this as a natural language search problem based on performance scores~\cite{zhou2023largelanguagemodelshumanlevel,yang2024largelanguagemodelsoptimizers}. 
To handle complex workflows, recent frameworks leverage reflexive feedback~\cite{shinn2023reflexionlanguageagentsverbal} or modular textual gradients~\cite{pryzant2023automaticpromptoptimizationgradient,khattab2024dspy} to optimize multi-stage pipelines. 
Furthermore, evolutionary strategies have been adopted to enhance exploration through population-based mutation and environmental feedback, such as EvoPrompt~\cite{guo2025evopromptconnectingllmsevolutionary}, GEPA~\cite{agrawal2025gepareflectivepromptevolution}, and SCOPE~\cite{pei2025scopepromptevolutionenhancing}. 
However, a critical limitation remains in \textit{context management}. Existing methods typically rely on scalar rewards or truncated summaries to fit context windows. In long-horizon agentic tasks, this compression obscures deep structural dependencies, hindering the diagnosis of complex, multi-step failures.


\subsection{Reflexive Agents and Harness Evolution}

Distinct from APO, reflexive and self-evolving agents improve agent behavior through feedback-driven revision, ranging from inference-time refinement~\cite{madaan2023selfrefineiterativerefinementselffeedback,sun2023adaplanneradaptiveplanningfeedback} and verifier-based reinforcement~\cite{jiang2025pagmultiturnreinforcedllm,zhou2025selfchallenginglanguagemodelagents} to code- and architecture-level evolution~\cite{novikov2025alphaevolvecodingagentscientific,sharma2025openevolve,hu2025automateddesignagenticsystems,zhai2025agentevolverefficientselfevolvingagent}. Recent harness-level systems, such as Hermes Agent Self-Evolution~\cite{nousresearch2026hermesagent}, further use optimizers like GEPA to evolve editable agent components. However, these methods still largely rely on unstructured feedback or trajectory-level reflection to decide what to modify. Without structured failure localization, they may repair downstream symptoms or update the wrong component. In contrast, \sysname{} performs dependency-guided diagnosis before optimization, producing localized repair signals that can transfer to skills and other tunable harness components.

\subsection{Agent Diagnostics and Credit Assignment}
The Credit Assignment Problem (CAP) fundamentally challenges the attribution of outcomes to specific decisions~\cite{pignatelli2024surveytemporalcreditassignment}. In Reinforcement Learning, this is traditionally addressed by mitigating reward sparsity via temporal redistribution mechanisms~\cite{harutyunyan2019hindsightcreditassignment,10.5555/3454287.3455502}. 
In modular agent systems, the focus shifts to Structural Credit Assignment (SCA), where recent frameworks utilize min-form bottleneck penalties~\cite{cheng2025stopsummationminformcredit} or generative critiques~\cite{xie2025capoenhancingllmreasoning} to refine reasoning chains. 
Despite these advancements, establishing explicit, causal-based attribution in long-horizon trajectories remains an open challenge.



\begin{figure*}[ht]
  \includegraphics[width=\linewidth]{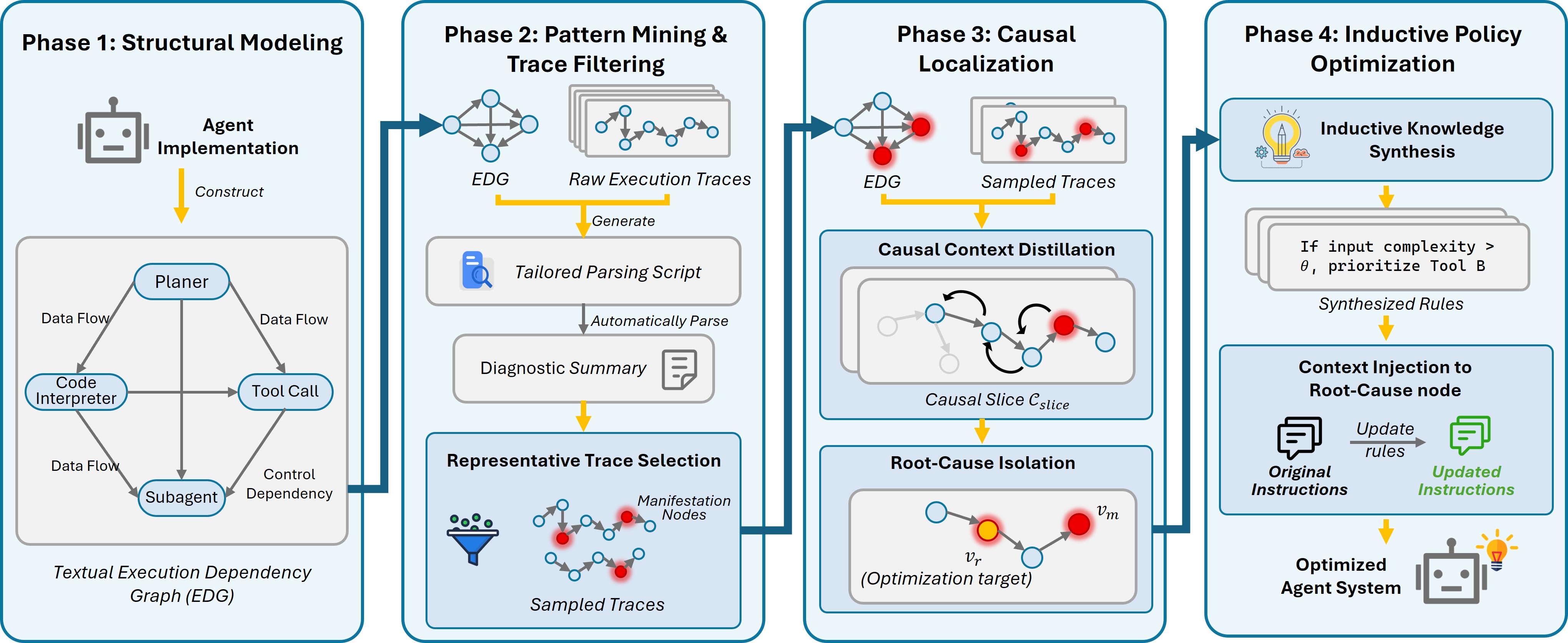}
  \caption {Illustration of the \textbf{\sysname{}} framework. The system optimizes long-horizon agents via four phases: (1) \textit{Structural Modeling}, which constructs execution topology; (2) \textit{Failure Pattern Mining and Trace Filtering}, which compresses failures into representative traces; (3) \textit{Causal Localization}, which performs causal context distillation and root cause isolation; and (4) \textit{Inductive Policy Optimization}, which injects heuristics into identified root cause modules.}
  \label{fig:pipeline}
\vskip -0.1in
\end{figure*}

\section{The \sysname{} Framework}

\subsection{System Overview}

\sysname{} is an advanced optimization framework designed to maximize the context signal-to-noise ratio (SNR) in the optimization of long-horizon agents. Directly placing massive traces into the agent optimizer's context window introduces substantial noise, diluting the LLM's reasoning signal and often trapping optimization in local optima. \sysname{} resolves this context-noise trade-off through a two-pronged approach: first, it performs trace filtering to retain only highly diverse, representative traces; second, it executes rigorous causal localization to explicitly strip away non-causal noise. By condensing large execution batches into high-quality diagnostic evidence, \sysname{} enables precise and stable optimization within finite context limits.

As illustrated in Figure~\ref{fig:pipeline}, the \sysname{} process operates through four integrated phases.
First, \textbf{\textit{Structural Modeling}} constructs a textual Execution Dependency Graph (EDG) to serve as the topological foundation for subsequent trace filtering. Second, \textbf{\textit{Failure Pattern Mining and Trace Filtering}} compresses massive, heterogeneous trace batches into a compact, diverse set of representative exemplars. Third, \textbf{\textit{Causal Localization}} distills the minimal causal context via backward slicing and isolates the true root cause node. Finally, \textbf{\textit{Inductive Policy Optimization}} converts this localized evidence into generalized, persistent prompt updates for the defective modules.

\subsection{Phase 1: Structural Modeling}

To enable structural analysis, \sysname{} leverages an LLM to read the agent's source implementation and construct a \textbf{textual Execution Dependency Graph (EDG)} $\mathcal{G}=(\mathcal{V}, E)$, which serves as the structural prior for the subsequent phases.
Specifically, \sysname{} parses the agent's codebase to identify the atomic functional modules (the vertex set $\mathcal{V}$). Simultaneously, it infers the textual edges $E$ by analyzing the artifacts these modules produce, consume, or utilize to dictate downstream execution paths. 

This construction yields a faithful, lightweight dependency map that captures two dependency types among modules: \textit{data dependencies}, where module $B$ consumes artifacts produced by module $A$ (e.g., intermediate plans, retrieved evidence, or execution outputs), and \textit{control dependencies}, where module $A$ governs the downstream execution path by deciding whether module $B$ is invoked or which tool/sub-agent is selected.
We provide the graph extraction procedure and detailed assessments in Appendix~\ref{app:textual_dependency_graph} and Tables~\ref{tab:graph_quality_models}--\ref{tab:graph_perturbation}, showing that the extracted graph is effective and remains useful for \sysname{} across different optimizer models.

\subsection{Phase 2: Failure Pattern Mining and Trace Filtering}
\label{sec:filtering}

Long-horizon agents often generate massive, heterogeneous failure traces during rollouts, making direct optimization inefficient and unstable. In this phase, \sysname{} compresses the raw trace set into a manageable size while rigorously preserving the diversity of failure modes. The pipeline operates in two steps: global diagnosis summarization followed by representative trace selection.

\noindent\textbf{Diagnosis Summarization.}
Instead of feeding raw traces directly to the LLM, \sysname{} generates a deterministic Python parser to extract key execution signals. To seamlessly adapt to varying agent logging formats, the system dynamically infers a trace schema from a single pilot trace, identifies structural delimiters (e.g., \textit{``Action:''}, \textit{``Observation:''}), and compiles a tailored parsing script on-the-fly. This parser then traverses the full trace set to record both the Global Outcome ($y \in \{Success, Failure\}$) and the Local Node Status (e.g., whether a specific module threw a runtime exception).

Upon processing the entire trace corpus, the system evaluates the aggregated dataset across two analytical dimensions:
(1) \textbf{\textit{Statistical Severity}}, which computes the conditional probability of a global task failure given a specific local node error, thereby identifying the most lethal bottlenecks; and
(2) \textbf{\textit{Structural Path Patterns}}, which tracks module invocation sequences to capture recurring pathological topologies. Specifically, this structural tracking detects anomalous behaviors such as infinite self-loops (e.g., a module repeatedly invoking the same tool without state progression) and dead-end trajectories (e.g., specific calling sequences that consistently lead to an abrupt timeout).
The output is a comprehensive diagnostic summary mapping these dominant failure patterns across the entire dataset, serving as the quantitative basis for trace selection (see Appendix~\ref{app:attribution_map_details} for an example).

\noindent\textbf{Representative Trace Selection.}
Because agent failures are heavily repetitive, \sysname{} applies diversity-based sampling to reduce trace volume. Based on the diagnostic summary, the framework clusters the trace data using both the \textit{statistical severity} of the errors and their \textit{structural path patterns}. It then selects a minimal exemplar set from each distinct cluster with a predefined size. This guarantees that the optimizer reviews a compact dataset covering all major failure modes with a controllable size, without being overwhelmed by redundancy.

\noindent\textbf{Fallback Mechanism:} In scenarios where the agent's trace lacks explicit node-level error signals (i.e., modules fail silently without raising exceptions), \sysname{} seamlessly adapts. It bypasses the local node status and clusters the traces relying on the global task outcome combined with the \textit{structural path patterns} of the module invocations. This ensures the selected exemplar set continues to rigorously capture distinct behavioral deviations.

\subsection{Phase 3: Causal Localization}
\label{sec:localization}

While the filtering phase successfully isolates a compact trace set, massive amounts of irrelevant information within these individual trajectories still make it difficult to locate the true root cause node. \sysname{} performs efficient causal localization to systematically pinpoint the core defect as an optimization target. 
To achieve this precision, the framework shifts the analytical focus: instead of just locating \textit{where} an error visibly manifests, it extracts \textit{what} exact context is causally responsible for it.
This naturally leads us to distinguish between the \textbf{Manifestation Node} $v_m$ (where the error explicitly surfaces, such as a code interpreter throwing a syntax exception; or the terminal node if no explicit error exists) and the \textbf{Root Cause Node} $v_r$ (the module harboring the actual logical defect). This phase bridges the gap between the two via a two-step process.

\noindent \textbf{Step 1: Causal Context Distillation via Backward Slicing.}
Given a representative trace and its manifestation node, \sysname{} extracts a minimal \textit{Causal Slice} containing only the historical steps that directly or indirectly influenced the failure. \sysname{} utilizes the EDG to traverse the trace in reverse. Starting from the failure state at $v_m$, the algorithm recursively traces the \textit{Data Dependency} and \textit{Control Dependency} edges backward.  
By retaining only steps that can reach the manifestation node through the dependency closure ($\rightarrow^*$), the system prunes nodes with no structural influence on the failure, thereby discarding mere temporal antecedents.

To illustrate, consider an agent executing multiple independent exploration branches. Because divergent reasoning paths are data-independent, an error manifesting on one specific branch possesses no causal link to the others. The backtracking mechanism bypasses nodes from those parallel, irrelevant explorations. The resulting high-SNR slice $\mathcal{C}_{slice}$ removes non-causal noise, capturing only the trajectory of the corrupted information flow.

\noindent \textbf{Step 2: Root Cause Isolation from the Causal Slice.}
Armed with the noise-free $\mathcal{C}_{slice}$, \sysname{} performs causal deduction to pinpoint  where the execution first deviated from the intended logic. The system traces the flow of corrupted context backward from the manifestation node. For instance, while a \textit{Code Interpreter} ($v_m$) might manifest the error by crashing, semantic reasoning over $\mathcal{C}_{slice}$ systematically reveals that an upstream \textit{Planner} generated a hallucinated parameter several steps earlier. Upon locating this originating logical defect, the system formally designates the module as the true \textbf{Root Cause Node} $v_r$.

\subsection{Phase 4: Inductive Policy Optimization}
\label{sec:repair}

By grouping the highly purified causal slices by their verified root cause nodes ($v_r$), the final phase transforms episodic errors into persistent policy upgrades, executing the actual optimization of the agent.

For each targeted module $v_r$, \sysname{} analyzes the aggregated causal slices to synthesize high-order generalized rules. Rather than translating recurring errors into instance-specific failure records (which risk overfitting), the optimizer performs inductive abstraction to formulate robust, reusable heuristics for future executions. \sysname{} then injects these newly discovered rules directly into the textual prompt policy of $v_r$. This guarantees that the targeted policy update resolves the exact structural deficiency at its source. The instruction templates are provided in Appendix~\ref{app:instructions}, and examples of synthesized heuristics are provided in Appendix~\ref{app:generalized_rules_details}.

\section{Experiments}

\begin{table*}[htbp]
  \caption{Performance of \sysname{} versus baselines on HotpotQA, WebArena, and VeruSAGE-Bench. HotpotQA is evaluated by Exact Match (EM), while WebArena and VeruSAGE-Bench are evaluated by Success Rate (SR). Parentheses in ALL indicate absolute gains over the Base Agent. Best results are highlighted in bold. TextGrad is omitted for VeruSAGE-Bench due to context-budget limits.
  }
  \label{tab:baseline}
  \begin{center}
    \begin{small}
      \setlength{\tabcolsep}{3pt}
      \begin{tabular*}{\textwidth}{@{\extracolsep{\fill}}lcccccc}
        \toprule
        \textbf{Prompt Setting}&\textbf{HotpotQA} &\multicolumn{5}{c}{\textbf{WebArena (SR)}}\\
        &(EM) &\textbf{Shopping}&\textbf{CMS}&\textbf{Reddit}&\textbf{GitLab}&\textbf{ALL}\\
        \midrule
        Base Agent & 37.0\% & 17.9\% & 10.8\% & 4.5\% & 7.3\% & 10.8\%\\
        Naive Few-shot& 60.0\% & 23.1\% & 13.5\% & 9.1\% & 7.3\% & 13.7\% (+2.9\%)\\
        Failure-Aware RAG & 57.7\%& 23.1\% & 16.2\% & 9.1\% & 7.3\% & 14.4\% (+3.6\%)\\
        Summary-based Selection & 64.3\% & 25.6\% & 10.8\% & 9.1\% & 12.2\% & 15.1\% (+4.3\%)\\
        Retrieval-based Selection & 46.0\% & 17.9\% & 13.5\% & 14.3\% & 12.2\% & 14.4\% (+3.6\%)\\
        TextGrad & 62.0\% & \textbf{30.8\%} & \textbf{16.2\%} & 9.1\% & 9.8\% & 17.3\% (+6.5\%)\\
        GEPA& 64.4\% & 23.1\% & 10.8\% & 27.3\% & 9.8\% & 16.5\% (+5.7\%)\\
        \textbf{\sysname{} (Ours)} & \textbf{68.5\%} & \textbf{30.8\%} & \textbf{16.2\%} & \textbf{36.4\%} & \textbf{17.1\%} & \textbf{23.7\% (+12.9\%)}\\
        \bottomrule
      \end{tabular*}

      \vspace{0.4em}

      \begin{tabular*}{\textwidth}{@{\extracolsep{\fill}}lcccccc}
        \toprule
        \textbf{Prompt Setting} & \multicolumn{6}{c}{\textbf{VeruSAGE-Bench (SR)}}\\
        &\textbf{IronKV}&\textbf{Memory Allocator}&\textbf{Node Replication}&\textbf{NRKernel}&\textbf{Storage}&\textbf{ALL}\\
        \midrule
        Base Agent & 41.6\%& 66.7\%& 80.0\%& 20.0\%& 46.2\%& 42.5\%\\
        Naive Few-shot & 37.5\%& 61.1\%& 80.0\%& 24.4\%& 30.8\%& 39.6\% (-2.9\%)\\
        Failure-Aware RAG & 37.5\%& 66.7\%& 70.0\%& 29.3\%& 38.5\%&42.5\% (+0.0\%)\\
        Summary-based Selection & 58.3\% & 61.1\% & 70.0\% & 29.3\% & 30.8\% & 45.3\% (+2.8\%) \\
        Retrieval-based Selection & 37.5\% & 66.7\% & 90.0\% & 22.0\% & 38.5\% & 41.5\% (-1.0\%) \\
        GEPA & 54.2\%& 61.1\%& 80.0\%& 29.3\%& 46.2\%& 47.2\% (+4.7\%)\\
        \textbf{\sysname{} (Ours)} & \textbf{62.5\%}& \textbf{88.9\%}& \textbf{100.0\%}& \textbf{31.7\%}& \textbf{61.5\%}& \textbf{58.5\% (+16.0\%)}\\
        \bottomrule
      \end{tabular*}
    \end{small}
  \end{center}
  \vskip -0.1in
\end{table*}

\subsection{Experimental Setup}

\noindent \textbf{Datasets.} We evaluate \sysname{} on three distinct benchmarks.
For general multi-hop reasoning, we employ \textbf{HotpotQA}~\cite{yang2018hotpotqa} in the multi-document setting, using a random $150/300$ train-test split.
For long-horizon agent optimization, we consider two complementary benchmarks. \textbf{WebArena}~\cite{zhou2024webarena} represents general-purpose web interaction, where agents complete realistic tasks across Shopping, CMS, Reddit, and GitLab tasks. \textbf{VeruSAGE-Bench}~\cite{verusage} represents domain-intensive formal verification, consisting of five real-world Rust projects~\cite{lattuada2024verus} with extreme context lengths (avg. 947 lines): IronKV, Memory Allocator, Node Replication, NRKernel, and Storage.
We enforce an $80\%/20\%$ split per project to test generalization. 

\noindent \textbf{Base Agent Implementation.} The optimization targets the set of instructions $\mathcal{P}$ that governs the modules in each workflow.
For HotpotQA, we use a DSPy-based~\cite{khattab2024dspy,agrawal2025gepareflectivepromptevolution} four-stage multi-hop reasoning workflow with 4 optimizable modules and \texttt{GPT-4o}~\cite{openai2024gpt4o}.
For WebArena, we use the original Chain-of-Thought (CoT) web-navigation agent with one tunable instruction and \texttt{GPT-4o} as the backbone.
For VeruSAGE-Bench, we use the hierarchical router-executor multi-agent framework~\cite{verusage} with 16 optimizable modules and \texttt{o4-mini}~\cite{openai2025o3o4mini} as the backbone. Each task is solved through an iterative verification-repair loop that runs for up to 20 repair attempts or 20 minutes. Detailed architectures and optimizable components are provided in Appendix~\ref{app:agent_details}.

\noindent \textbf{Baselines.} We compare \sysname{} against the unoptimized Base Agent, and three categories of optimization baselines. For \textit{static strategies}, we utilize Naive Few-shot~\cite{brown2020languagemodelsfewshotlearners}, and a Failure-Aware RAG baseline~\cite{lewis2021retrievalaugmentedgenerationknowledgeintensivenlp} that retrieves traces via a BM25 index~\cite{robertson2009probabilistic} and appends them as instruction references.
For \textit{automated optimizers}, we select two representative methods. TextGrad~\cite{yuksekgonul2024textgrad} represents the gradient-based paradigm, which feeds the full execution trace to the optimizer to compute error attributions, and GEPA~\cite{agrawal2025gepareflectivepromptevolution} represents the evolutionary paradigm, which typically applies context truncation to fit recent history into the optimizer window.
For \textit{context management baselines}, we include two common trace compression strategies for handling extremely long traces during optimization: Summary-based and Retrieval-based optimization. Both use the same optimizer as \sysname{} but construct the optimization context without causal localization: the former summarizes the full trace before inductive abstraction, while the latter retrieves the most relevant steps as optimizer input.
\textit{Note: TextGrad is excluded from VeruSAGE-Bench due to prohibitive context costs (>100k tokens).}

\noindent \textbf{Metrics \& Configuration.} We report Exact Match (EM) for HotpotQA and Success Rate (SR) for WebArena and VeruSAGE-Bench. Optimization cost is reported in USD for one full optimization run, estimated from token usage under API pricing.
For the optimizer, we employ \texttt{Claude Sonnet 4.5} (temp=1.0) as the meta-controller for all methods to ensure fairness, with a bottleneck threshold $k=5$ and exemplar count $s=5$.

\subsection{Main Results}
\label{sec:main_results}

Table~\ref{tab:baseline} summarizes the performance of \sysname{} across benchmarks that cover both general reasoning and complex long-horizon agentic tasks. Across these settings, \sysname{} achieves the strongest overall performance compared to all baselines. The pattern across baseline categories helps explain where the gains of \sysname{} come from. Static strategies can provide additional examples or failure cases,  but they fail to identify which part of the current behavior should be revised. Automated optimizers such as TextGrad and GEPA still rely on provided context quality: using full trajectories can exceed the context budget or dilute local signals, whereas simple truncation or relevance-based slicing may miss the upstream root cause. Context management methods attempt to reduce the trace context, but summaries may discard fine-grained failure signals, while retrieval can be misled by semantic similarity. In contrast, \sysname{} addresses this context-noise trade-off through a dependency-guided optimization framework that provides localized yet context-aware signals, enabling the optimizer to focus on root causes rather than noisy downstream symptoms and thereby achieve the strongest overall performance across benchmarks.

Beyond the overall comparison, \sysname{} also shows strong adaptability to two optimization needs in agentic tasks. On WebArena, the key difference lies in the task-specific demand for optimization signals: Shopping and CMS require more workflow guidance, while Reddit and GitLab require more failure-aware rules. In the former domains, \sysname{} remains competitive with the strongest baselines by preserving useful procedural context. In the latter domains, \sysname{} achieves clearer improvements, suggesting that \sysname{} leverages both workflow-level patterns and failure-aware optimization signals.
The advantage of \sysname{} becomes more pronounced on the longer-horizon and more domain-intensive VeruSAGE-Bench, where \sysname{} achieves the highest average success rate of \textbf{58.5\%}, outperforming the base agent by \textbf{+16.0\%} and the strongest baseline, GEPA, by \textbf{+11.3\%}. These results demonstrate the effectiveness of \sysname{} especially in complex agentic settings, where failures may propagate across multiple components and require root-cause localization rather than shallow reuse of prior traces.

\subsection{Optimization Scalability and Cost Analysis}
\label{sec:efficiency}

\begin{figure}[t]
  \begin{center}
    \centerline{\includegraphics[width=\columnwidth]{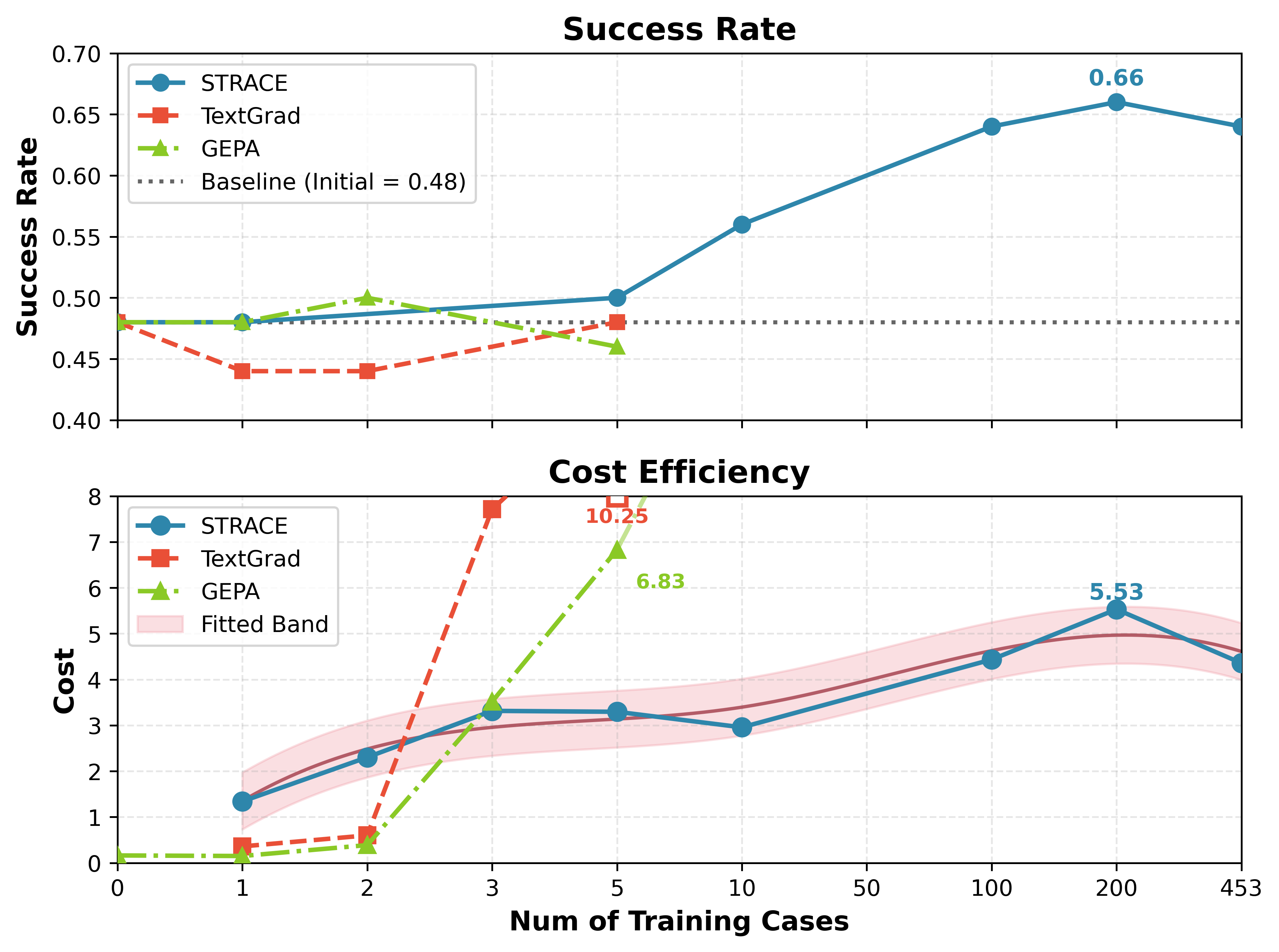}}
    \caption{
    Success rate on 50 independent tasks and optimization cost for \sysname{}, GEPA, and TextGrad full-trace prompt optimization, as the number of training cases scales from 1 to 453.}
    \label{fig:tradeoff}
  \end{center}
  \vskip -0.3in
\end{figure}

Long-horizon prompt optimization must balance performance with cost, as naively increasing context quickly becomes prohibitive. Figure~\ref{fig:tradeoff} plots the success rate and optimization cost against the number of training trajectories. For a fair comparison, all methods are optimized on the same trajectory subset at each scale point. We include two representative baselines: \textbf{TextGrad} (full-trace optimization) and \textbf{GEPA} (current node slicing and batch optimization).

As the training set scales, \sysname{} achieves the most favorable cost--performance trend. While TextGrad incurs rapidly increasing costs due to full-trace processing and GEPA reduces cost by restricting optimization to local-node batches, \sysname{} maintains both efficiency and effectiveness. This favorable trade-off is enabled by the \textit{Statistical Bottleneck Diagnosis} module, which summarizes dataset-level failure patterns and selects only a small set of representative, high-value exemplars for optimization. As a result, even when scaling to the full set of \textbf{453} trajectories, \sysname{} controls context growth while preserving the failure evidence needed for effective repair, yielding the strongest overall cost--performance trend among all methods.

\subsection{Case Study: Tracing Symptoms to Root Causes}
\label{sec:case_study}

\begin{figure}[t]
  \begin{center}
    \centerline{\includegraphics[width=\columnwidth]{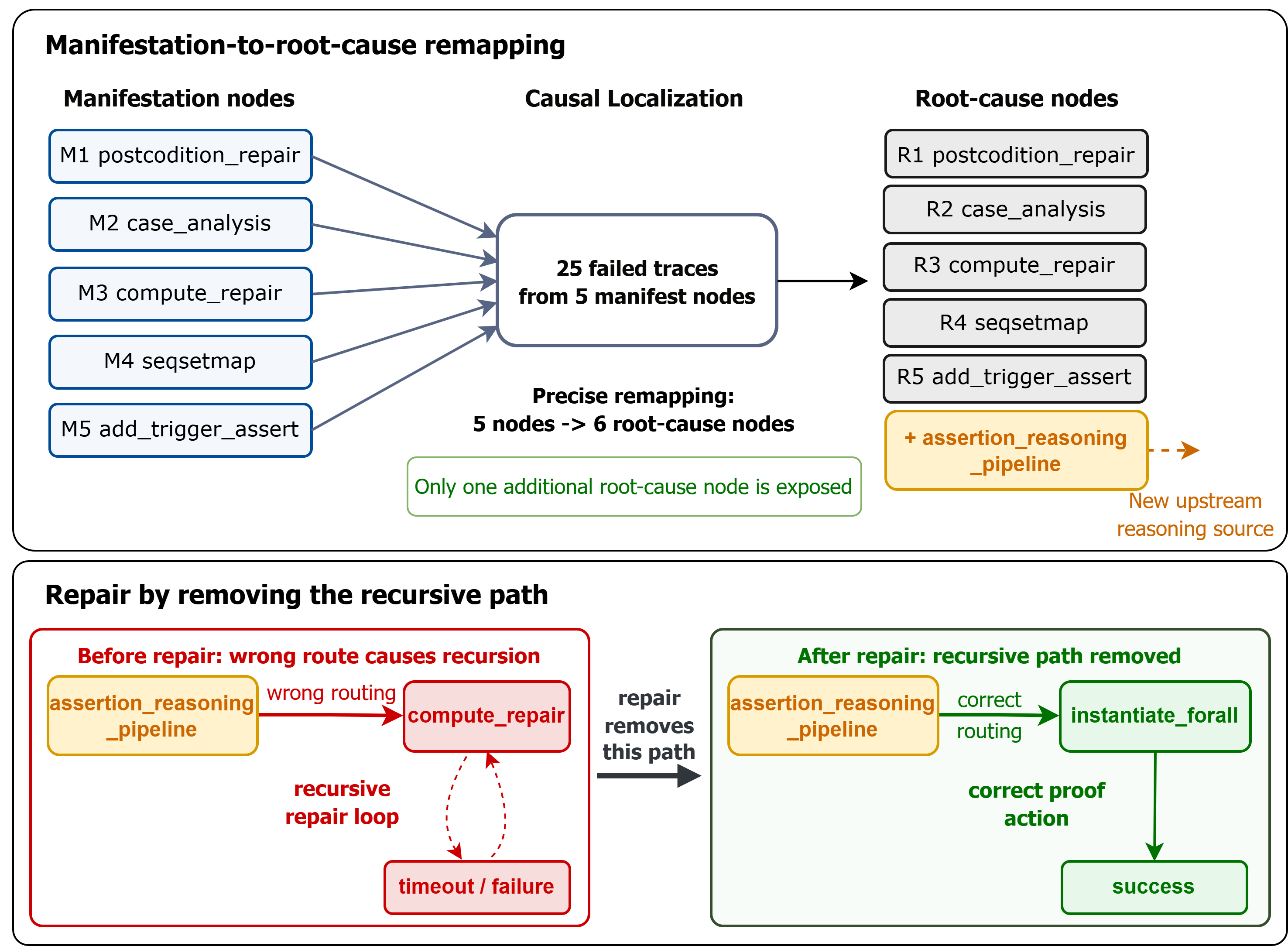}}
    \caption{\textbf{Distinguishing Symptoms from Root Causes.} \textbf{Top:} Causal localization remaps 25 traces from 5 manifest nodes to 6 targets, exposing an upstream source. \textbf{Bottom:} A \texttt{compute\_repair} loop is traced to \texttt{assertion\_reasoning\_pipeline}, enabling a targeted policy optimization to bypass the recursive path.}
    \label{fig:case_study}
  \end{center}
  \vskip -0.3in
\end{figure}

To empirically validate the distinction between \textit{symptom} ($v_m$) and \textit{cause} ($v_{r}$), we present two analyses from the VeruSAGE-Bench optimization process in Figure~\ref{fig:case_study}. These cases illustrate how root cause isolation redirects optimization from surface-level failures to upstream logical defects.

\noindent \textbf{Re-mapping Target Nodes.} 
As formalized in Section \ref{sec:localization}, 
effective optimization requires targeting the root-cause node rather than the manifestation node.
In the 200-case setting, \sysname{} selects 5 high-impact manifest modules and 5 representative traces for each, yielding 25 traces in total. Without causal localization, these traces would be attributed to the modules where failures surface naturally. However, by applying root cause isolation, 12 of the 25 traces are re-mapped to upstream root-cause actions, expanding the optimization targets to 6 distinct nodes, with an additional upstream ``\texttt{assertion\_reasoning\_pipeline}'' module.
This redistribution confirms that relying solely on crash location ($v_m$) can lead to ``optimizing the wrong target'' for a subset of failures, whereas \sysname{} disperses the diagnosis to the true logical origins.

\noindent \textbf{Breaking the Loop.}
We further examine a representative IronKV failure. The agent repeatedly invokes a downstream repair module until timeout, making the visible failure appear local to that module. Causal slicing instead traces the loop back to an earlier routing error: the upstream controller dispatched the task to an unsuitable repair path. By repairing this root-cause decision, \sysname{} breaks the loop and enables successful verification.

\subsection{Ablation Study}


\begin{table}[t]
  \caption{Ablation results with Success Rate and stage-wise cost breakdown.}
  \label{tab:ablation}
  \begin{center}
    \begin{small}
      \setlength{\tabcolsep}{2pt}
      \begin{tabular*}{\columnwidth}{@{\extracolsep{\fill}}lcc}
        \toprule
        Ablation Method&SR & Overall Cost (\$)\\
        \midrule
        \textbf{\sysname{}}&\textbf{56\%}&\textbf{2.96}\\
        w/o Structural Modeling& 48\%&5.10\\
        w/o Trace Filtering& 46\%&8.45\\
        w/o Causal Localization (Current)& 54\%&2.88\\
        w/o Causal Localization (Full)& 54\%&5.93\\
        \bottomrule
      \end{tabular*}
    \end{small}
  \end{center}
  \vskip -0.2in
\end{table}

We ablate \sysname{} to isolate the contributions of structure modeling, trace filtering, and causal localization. Each variant is optimized under the same protocol with the same training traces. 
Table~\ref{tab:ablation} reports the resulting Success Rate and overall cost. Removing structure modeling leads to lower effectiveness and higher cost, suggesting that the dependency graph provides a useful prior for organizing trace failures and narrowing the search space before optimization. Removing trace filtering further degrades both performance and efficiency, as the optimizer must process more redundant or low-value traces rather than a compact set of representative failures. The causal localization variants reveal the context-noise dilemma: using only the current node may miss upstream causes, while using the full trace introduces unnecessary noise and cost. In contrast, \sysname{} combines these components to preserve root-cause evidence while keeping the optimization context bounded. Detailed cost-component analyses are deferred to Appendix~\ref{app:cost_analysis}.

\section{Conclusions}


We introduce a novel method \sysname{} for long-horizon agent optimization under the context-noise trade-off. By combining failure pattern mining with causal localization, \sysname{} breaks this dilemma, enabling targeted policy optimization that is both cost-efficient and high-performing. Results on HotpotQA, WebArena, and VeruSAGE-Bench show consistent gains in success rate and efficiency, outperforming all baselines and demonstrating the effectiveness of our method. Ablation studies and analysis further validate the contribution of each phase. Importantly, \sysname{} can be readily transferred to harness engineering, where its localized diagnoses can guide updates to tunable agent components such as skills, tool-use policies, and other editable harness elements. This transferability makes \sysname{} a flexible framework for improving complex long-horizon agents.

\section*{Limitations}

A main limitation of \sysname{} is that it assumes sufficient visibility into the target agent system. The dependency prior is constructed from codebase-level or harness-level artifacts, including agent definitions, tool interfaces, configuration files, prompts, skills, and execution logs. Therefore, \sysname{} is most directly applicable to systems where the optimizer can inspect the implementation or harness, rather than fully black-box agents with hidden components and control flow. Notably, this does not require access to model weights, but it does require enough system-level visibility to identify the components involved in long-horizon failures. Extending \sysname{} to purely trace-only or black-box settings remains future work.

\section*{Ethical considerations}

In this study, all datasets, models, and software artifacts used in our experiments are publicly available, and we follow their respective licenses or terms of use. The use of these resources is strictly confined to academic research and benchmarking purposes. We do not collect new human-subject data, and all reported results are presented in aggregate form.
While \sysname{} is designed to improve the efficiency and reliability of long-horizon agent optimization, we acknowledge that stronger agent optimization methods could be misused if applied to unsafe or malicious agent systems. In addition, execution traces collected from real deployments may contain sensitive information. For practical deployment, we recommend that traces be anonymized or access-controlled, and that optimized prompts, skills, or harness components be reviewed before being applied to high-stakes or user-facing systems.

\bibliography{custom}

\clearpage

\section*{Appendix}
\label{sec:appendix}
\appendix

\section{Agent Implementation Details}
\label{app:agent_details}

Here we provide detailed specifications of the base agents used in our experiments, including their architectural workflows and the specific sub-agents targeted for optimization.
\subsection{HotpotQA Agent Workflow}
Following the setup in GEPA~\cite{agrawal2025gepareflectivepromptevolution}~\footnote{\url{https://github.com/gepa-ai/gepa-artifact}}, our HotpotQA agent consists of four sequential sub-agents designed to handle multi-hop reasoning. The workflow proceeds as follows:

\begin{enumerate}[itemsep=0pt, parsep=0pt, topsep=2pt]
    \item \textbf{Summarize1:} This module processes the initial question and the paragraphs retrieved in the first hop. Its goal is to extract relevant information and form a preliminary context.
    \item \textbf{Create Query Hop2:} Based on the initial summary and the original question, this module generates a refined search query to retrieve missing evidence required for the second hop.
    \item \textbf{Summarize2:} This module integrates the newly retrieved passages from the second hop with the prior context, synthesizing a comprehensive summary that consolidates evidence from both retrieval rounds.
    \item \textbf{Final Answer:} Finally, this module reasons over the aggregated summaries to produce the final answer string (and supporting facts) for the multi-hop question.
\end{enumerate}

During optimization, \sysname{} optimizes the system prompts for all four modules simultaneously.

\subsection{WebArena Agent Workflow}

For WebArena, we followed the original WebArena Chain-of-Thought agent design~\cite{zhou2024webarena}~\footnote{\url{https://github.com/web-arena-x/webarena}}. WebArena agent is a single prompt-based web navigation agent rather than a multi-agent system. At each step, the agent observes the current webpage through a textual accessibility tree, together with the current URL, the user instruction, and the previous action. A language model then reasons over this information and predicts the next browser action.

The interaction loop proceeds as follows:

\begin{enumerate}[itemsep=0pt, parsep=0pt, topsep=2pt]
    \item \textbf{Observe:} The browser environment converts the current webpage into an accessibility-tree representation, exposing interactive elements such as buttons, links, text boxes, and menus.
    \item \textbf{Reason and Act:} Given the task objective and the current observation, the Chain-of-Thought prompt guides the language model to reason about the page state and output one concrete action, such as clicking an element, typing into a field, scrolling, navigating to a URL, or stopping with a final answer.
    \item \textbf{Execute:} The predicted action is executed in the browser environment, producing a new webpage state.
    \item \textbf{Repeat:} The agent repeats this observe--reason--act cycle until it issues a stop action or reaches the maximum interaction budget.
\end{enumerate}

During optimization, \sysname{} targets the prompt of this single WebArena agent. Since the baseline does not contain explicit sub-agents, planners, or external memory modules, the optimized component is the agent's instruction prompt, which defines its action space, reasoning style, and few-shot behavior examples.

\subsection{VeruSAGE Agent Workflow}
For VeruSAGE-Bench, we utilize the VeruSAGE framework~\cite{verusage}~\footnote{\url{https://github.com/microsoft/verus-proof-synthesis}}, which employs a more complex, iterative multi-agent architecture to repair verification failures in Rust code.

\textbf{Iterative Repair Loop.}

The agent solves each task through a verification-driven loop. In each iteration:
\begin{enumerate}[itemsep=0pt, parsep=0pt, topsep=2pt]
    \item The current code is submitted to the Verus verifier.
    \item If verification fails, the \textit{Error Analyzer} parses the error message.
    \item A \textit{Router} (or High-level Planner) selects the most appropriate repair action from a pool of 35 available actions based on the error type and code context.
    \item The selected \textit{Sub-agent} (Executor) generates a specific patch (e.g., adding a lemma function, enabling a specific prover mode, applying divide-and-conquer for a specific property to be proved).
    \item The patch is applied, and the loop repeats.
\end{enumerate}
The process terminates when the code verifies successfully or when the budget (20 attempts or 20 minutes) is exhausted.

\textbf{Sub-agent Structure.}
The framework comprises 15 specialized sub-agents covering various aspects of formal verification, including but not limited to:
\begin{itemize}[itemsep=0pt, parsep=0pt, topsep=2pt]
    \item \textit{Invariant Manager:} For analyzing and fixing loop invariants.
    \item \textit{Proof Strategy Applicator:} For conducting proof strategies like divide-and-conquer (i.e., case analysis) and induction. 
    \item \textit{Prover Mode Activator:} For enabling special prover modes, like bit-vector mode, non-linear arithmetic mode, etc. 
    \item \textit{Syntax Fixer:} For correcting Rust-specific compilation errors.
    \item \textit{Standard Lemma Guider:} For suggesting what lemmas inside the Verus standard library might be useful.
\end{itemize}

\textbf{Optimization Scope.}
Due to the domain-intensive nature of Verus, generic prompts often fail to capture the nuances of formal reasoning. \sysname{} targets the system prompts of the \textit{Router} and the top-k most frequently triggered \textit{Executors}, injecting synthesized domain heuristics (e.g., specific strategies for handling "nonlinear arithmetic" errors) directly into their context windows.

\section{\sysname{} Implementation Details}

\subsection{Architecture and Context Management}

\sysname{} is implemented on top of the Claude-Agent-SDK~\cite{anthropic2025claudeagentsdk} as a four-phase agent workflow. From an architectural perspective, the system separates the optimization process into four sequential phases, while adopting different context management strategies for different stages. Phases 1--3 are executed within a shared context window. This design is motivated by the strong continuity across these stages: all three phases rely on the same agent background, and they also need to share the textual dependency graph inferred during structural analysis. As a result, there is a natural flow of information across the first three phases, where intermediate outputs produced in one phase directly serve as inputs to the next. In contrast, the Phase 4 is executed in an isolated context window supported by the reason that \sysname{} has already identified the true root-cause node and produced a concise explanation of why this node is the root cause within the slicing chain by the end of Phase 3. Therefore, Phase 4 no longer needs access to the full intermediate context from the earlier stages. Instead, it only needs the local context around the root-cause node. Separating Phase 4 from the earlier shared context thus reduces unnecessary context overhead and keeps the final revision step focused and modular.

\subsection{Textual Dependency Graph Construction}
\label{app:textual_dependency_graph}

In our experiments, \sysname{} instantiates Phase~1 with an optimizer agent built on the Claude Agent SDK under a dedicated textual dependency-graph construction instruction. The agent first inspects the target repository for available background artifacts, including README files, documentation, configuration files, agent definitions, and relevant source code. 
It then identifies the major modules in the system, including active modules (e.g., decision makers and executors), passive shared-state modules (e.g., memory or state managers), and the key artifacts each module produces or consumes. Based on this repository-level inspection, Phase~1 infers two types of dependency priors: data dependencies and control dependencies.
We refer to this textual dependency graph as the dependency prior and denote it by $\mathcal{G}$. Notably, the term ``graph'' reflects its node--edge structure, but its implementation is a text edge list rather than a learned graph object or a static-program-analysis graph. Each edge is a textually specified relation between two components, together with its dependency type and brief evidence from repository artifacts. Later diagnosis and slicing phases consume this textual graph as structural guidance.

This construction is framework-agnostic: the analyzed agent system may be implemented in different frameworks or coding styles, and \sysname{} does not require framework-specific APIs for tools or agents. Instead, it recovers dependency priors from repository-level artifacts such as code, configuration, and agent definitions.

To further analyze the role of graph quality in \sysname{}, we report two complementary studies: an expert-reviewed comparison of repository-constructed graphs across models and a graph perturbation experiment. These studies address two questions: whether the dependency prior can be reliably constructed by different optimizer models, and whether \sysname{} remains effective when this prior is noisy.


\begin{table*}[t]
  \caption{Expert-reviewed quality assessment of dependency graphs constructed from the repository by different models.}
  \label{tab:graph_quality_models}
  \begin{center}
    \begin{small}
      \setlength{\tabcolsep}{5pt}
      \resizebox{0.99\linewidth}{!}{%
      \begin{tabular}{lp{0.67\textwidth}}
        \toprule
        Model (constructed from repository) & Graph Quality Assessment (expert-reviewed) \\
        \midrule
        Claude Sonnet 4.5~\cite{anthropic2025sonnet45} & Highest-quality graph, preserves the core dependency structure with appropriate granularity.\\
        Claude Haiku 4.5~\cite{anthropic2025haiku45} & Preserves the core dependency structure, with some additional infrastructure-level edges.\\
        GPT-5.1 Codex Mini~\cite{openai2025gpt5} & Preserves the core dependency structure at a finer granularity.\\
        DeepSeek-V3.1~\cite{deepseek_v31_2025} & Preserves the core dependency structure, with additional infrastructure-level details.\\
        MiniMax-M2.5~\cite{minimax_m25_2026} & Preserves the core dependency structure, with about 10\% additional noisy edges.\\
        \bottomrule
      \end{tabular}%
      }
    \end{small}
  \end{center}
\end{table*}

\textbf{RQ1}: Can different optimizer models construct a reliable dependency prior from the repository?


Table~\ref{tab:graph_quality_models} answers RQ1 by comparing dependency graphs constructed from the same repository using different models. The expert review shows that all models recover the core dependency structure needed by \sysname{}. Their differences mainly appear in granularity and the amount of extra infrastructure-level detail: stronger models produce cleaner and more appropriately scoped graphs, while weaker models may introduce additional noisy edges. Importantly, even the weakest graph still preserves the core dependencies and introduces only about 10\% noisy edges, suggesting that the dependency-prior construction is reasonably stable across optimizer backbones.

\textbf{RQ2}: How sensitive is \sysname{} to noise in the dependency prior?


To answer RQ2, we conducted controlled perturbation experiments on the dependency prior, as shown in Table~\ref{tab:graph_perturbation}. We randomly add or remove 10\% or 25\% of edges between graph nodes to simulate missing dependencies or additional noisy dependencies. Removing the dependency graph entirely causes performance to fall back to the base-agent level (48\%), confirming that the structural prior is essential for effective optimization.

In contrast, \sysname{} remains stable when the graph is only partially noisy: removing 25\% of edges reduces success rate only slightly from 56\% to 54\%, while adding 25\% of edges preserves the same 56\% success rate. At the smaller 10\% perturbation level, performance is preserved or even improved. These results indicate that \sysname{} benefits substantially from having a dependency prior, but does not require this prior to be perfectly specified.

\begin{table}[t]
  \caption{Impact of dependency-graph perturbations on Success Rate.}
  \label{tab:graph_perturbation}
  \begin{center}
    \begin{small}
      \setlength{\tabcolsep}{8pt}
      \begin{tabular*}{\columnwidth}{@{\extracolsep{\fill}}lc}
        \toprule
        Graph Settings & SR \\
        \midrule
        Base Agent (No optimization) & 48\%\\
        No dependency graph & 48\%\\
        Remove 25\% edges & 54\%\\
        Add 25\% edges & 56\%\\
        Remove 10\% edges & 56\%\\
        Add 10\% edges & 62\%\\
        \textbf{STRACE} & \textbf{56\%}\\
        \bottomrule
      \end{tabular*}
    \end{small}
  \end{center}
\end{table}

\subsection{Visualization of Failure Pattern Mining}
\label{app:attribution_map_details}

\begin{figure*}[t]
\centering
  \includegraphics[width=\linewidth]{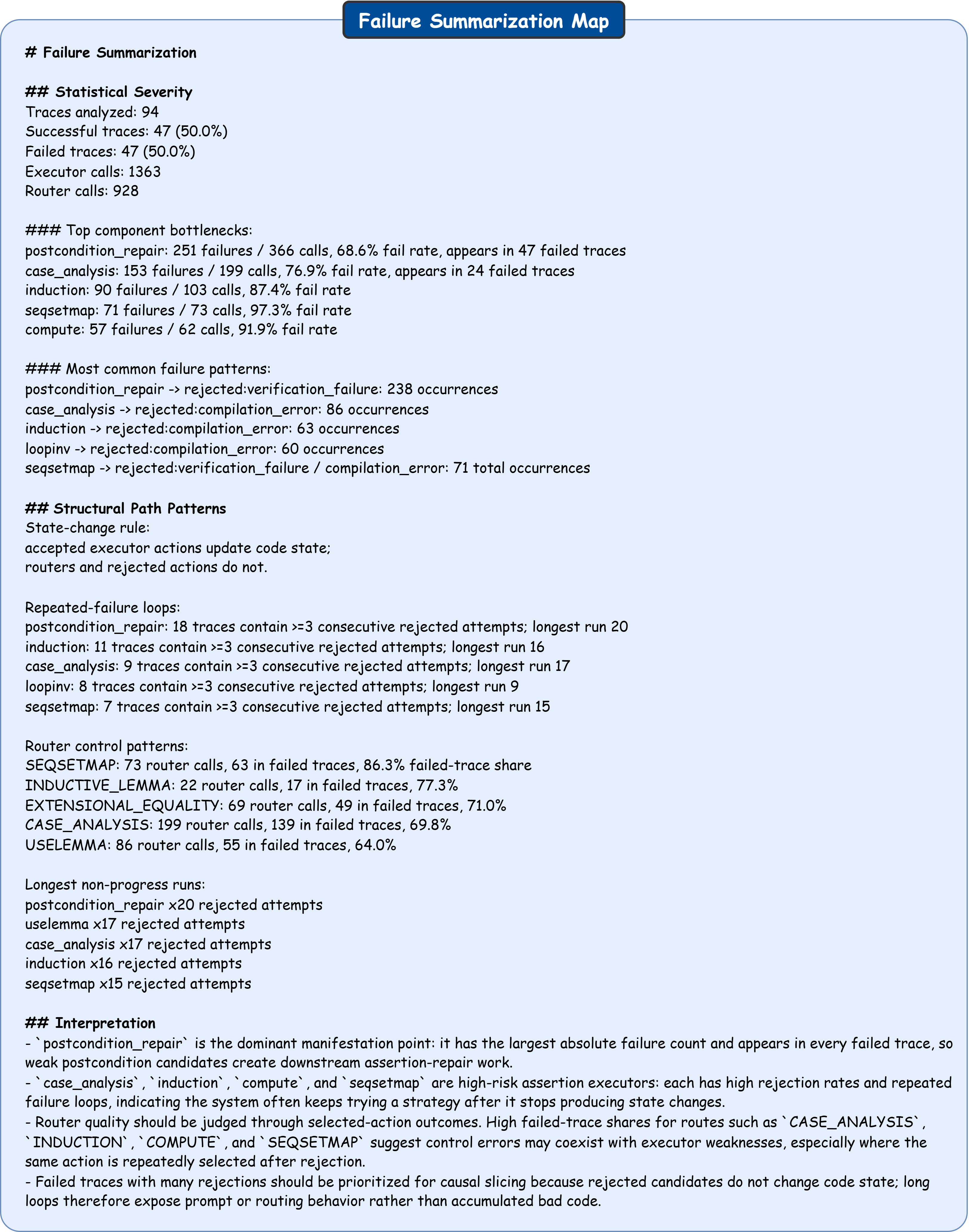}
  \caption {
    Visualization of a \textbf{Failure Summarization Map} generated during the optimization of the \textit{IronKV} project.
  \sysname{} first parses raw traces into structured execution records containing the global task outcome, local node status, and
  module invocation sequence.
  The map then summarizes the trace corpus along two analytical dimensions:
  \textit{Statistical Severity}, which estimates the conditional probability of global failure given each local node error, and
  \textit{Structural Path Patterns}, which identifies recurring pathological trajectories such as self-loops, oscillations, and
  dead ends.
  These aggregated diagnostics serve as the quantitative basis for selecting high-value traces for causal localization.
    }
  \label{fig:failure_attribution_map}
\end{figure*}


Figure~\ref{fig:failure_attribution_map} provides a concrete example of the diagnostic summary produced by the failure pattern mining stage.
Rather than feeding raw traces directly into the optimizer, \sysname{} first converts heterogeneous logs into structured records and aggregates them into dataset-level failure patterns.
This summary supports trace filtering through two complementary signals:

\begin{itemize}[itemsep=0pt, parsep=0pt, topsep=2pt]




\item \textbf{(a) Statistical Severity:}
The summary estimates the conditional failure probability $P(\text{Task Fail} \mid v_i \text{ Fail})$, where $v_i$ denotes a local node error extracted from the structured trace record.
This identifies local failures that are strongly associated with global task failure, allowing \sysname{} to prioritize systemic bottlenecks over low-impact or stochastic errors.

\item \textbf{(b) Structural Path Patterns:}
The summary also tracks module invocation sequences across traces to reveal recurring pathological trajectories.
Patterns such as self-loops, oscillations, and dead ends indicate failure modes that cannot be captured by node-level statistics alone, and help select traces that preserve diverse structural failure behaviors.

\end{itemize}

\begin{table*}[h]
  \caption{Ablation results with stage-wise cost breakdown.}
  \label{tab:ablation_cost}
  \begin{center}
    \begin{small}
        \resizebox{0.99\linewidth}{!}{%
        \begin{tabular}{c|ccccc}
        \toprule
          & Cost at Phase1&Cost at Phase2&Cost at Phase3&Cost at Phase4&Overall Cost\\
          \midrule
          STRACE&0.1135&0.1861&0.8515&1.8052&2.9563\\
 w/o Structural Modeling& - & 1.0099 & 2.2212 & 1.8734 & 5.1045\\
 w/o Trace Filtering& 0.1667& -& 0.6802& 7.602&8.4489\\
 w/o Causal Localization (Current Node)& 0.2956& 0.5320& -& 2.0486&2.8762\\
 w/o Causal Localization (Full Trace)& 0.2500& 0.3597& -& 5.3179&5.9276\\
 \bottomrule
 \end{tabular}%
        }
    \end{small}
  \end{center}
\end{table*}

\subsection{Instructions of Phases}
\label{app:instructions}

\begin{figure*}[h]
  \centering
  \includegraphics[width=\textwidth]{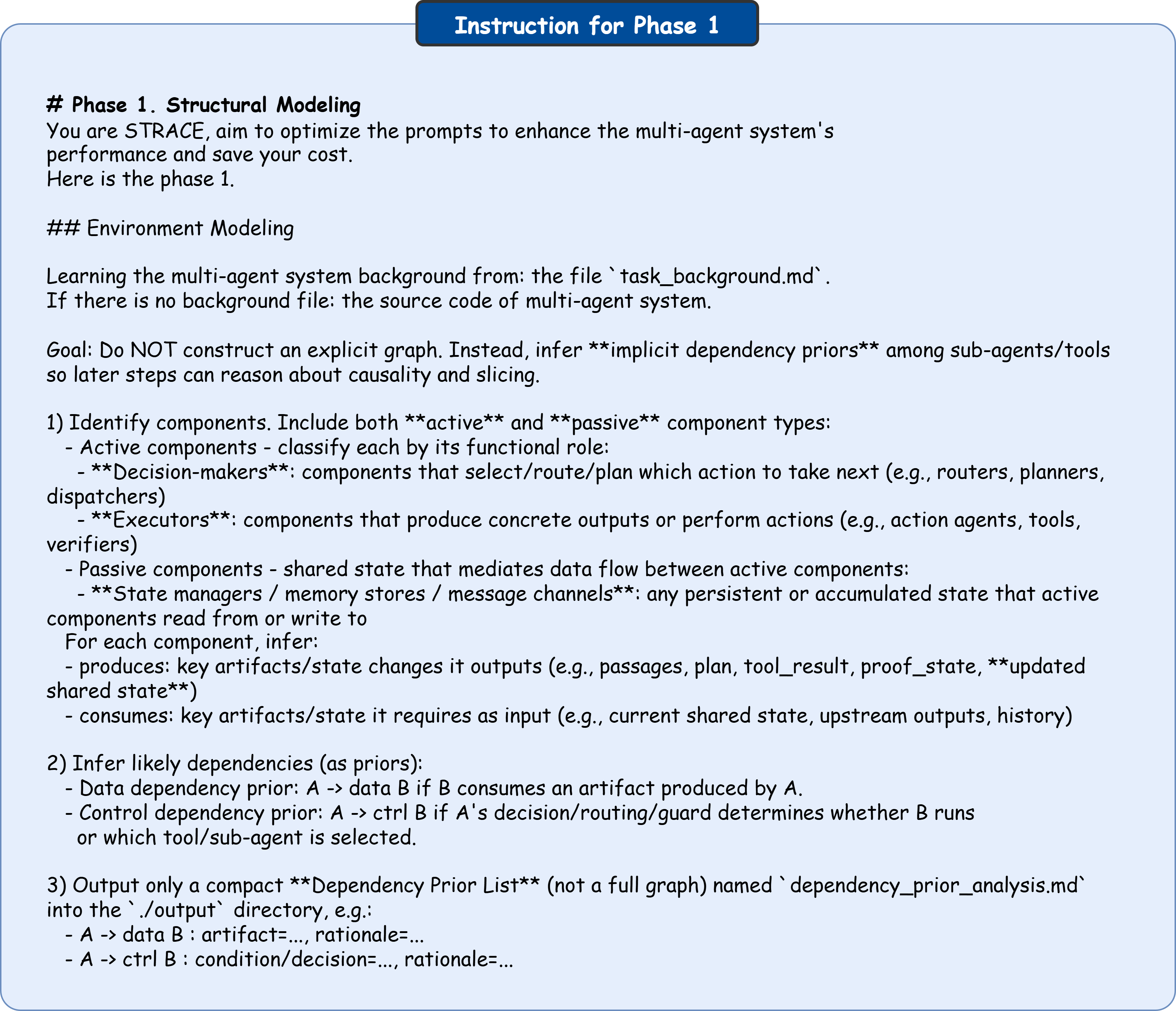}
  \caption{Instruction for Phase 1--Structural Modeling.}
  \label{fig:phase1_instruction}
  
\end{figure*}

\begin{figure*}[h]
  \centering
  \includegraphics[width=\textwidth]{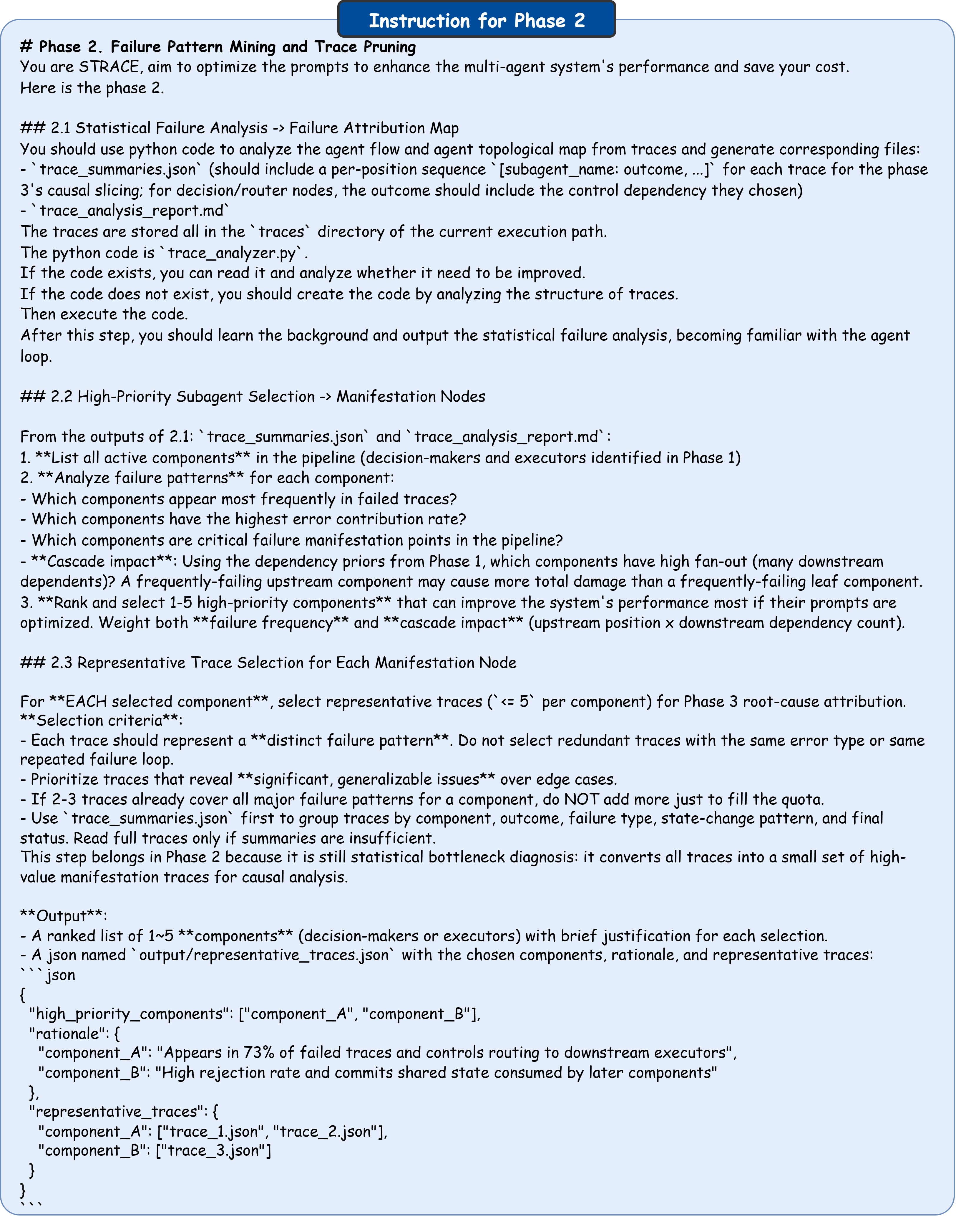}
  \caption{Instruction for Phase 2 -- Failure Pattern Mining and Trace Filtering.}
  \label{fig:phase2_instruction}
  
\end{figure*}

\begin{figure*}[h]
  \centering
  \includegraphics[width=\textwidth]{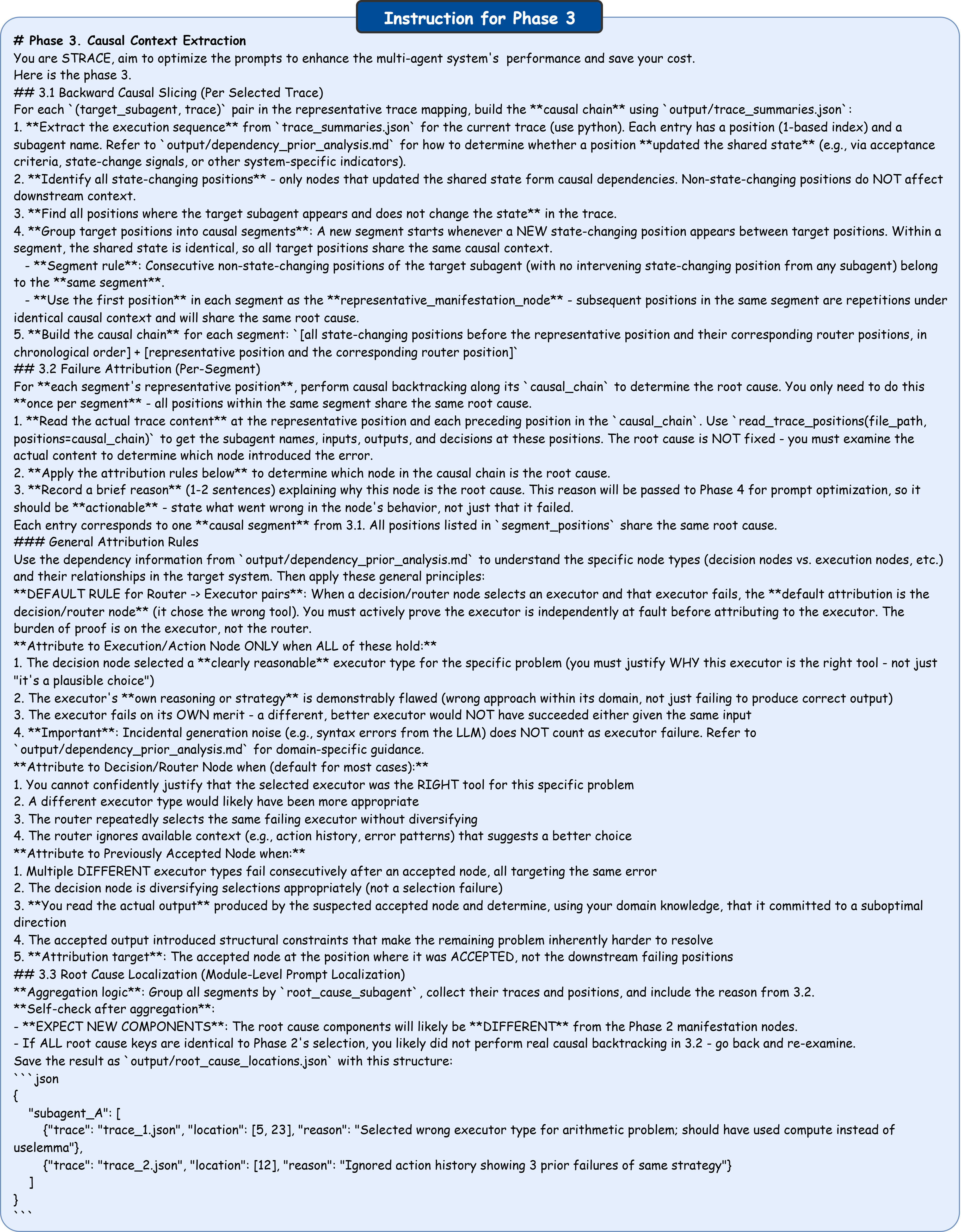}
  \caption{Instruction for Phase 3 -- Causal Localization.}
  \label{fig:phase3_instruction}
  
\end{figure*}

\begin{figure*}[h]
  \centering
  \includegraphics[width=\textwidth]{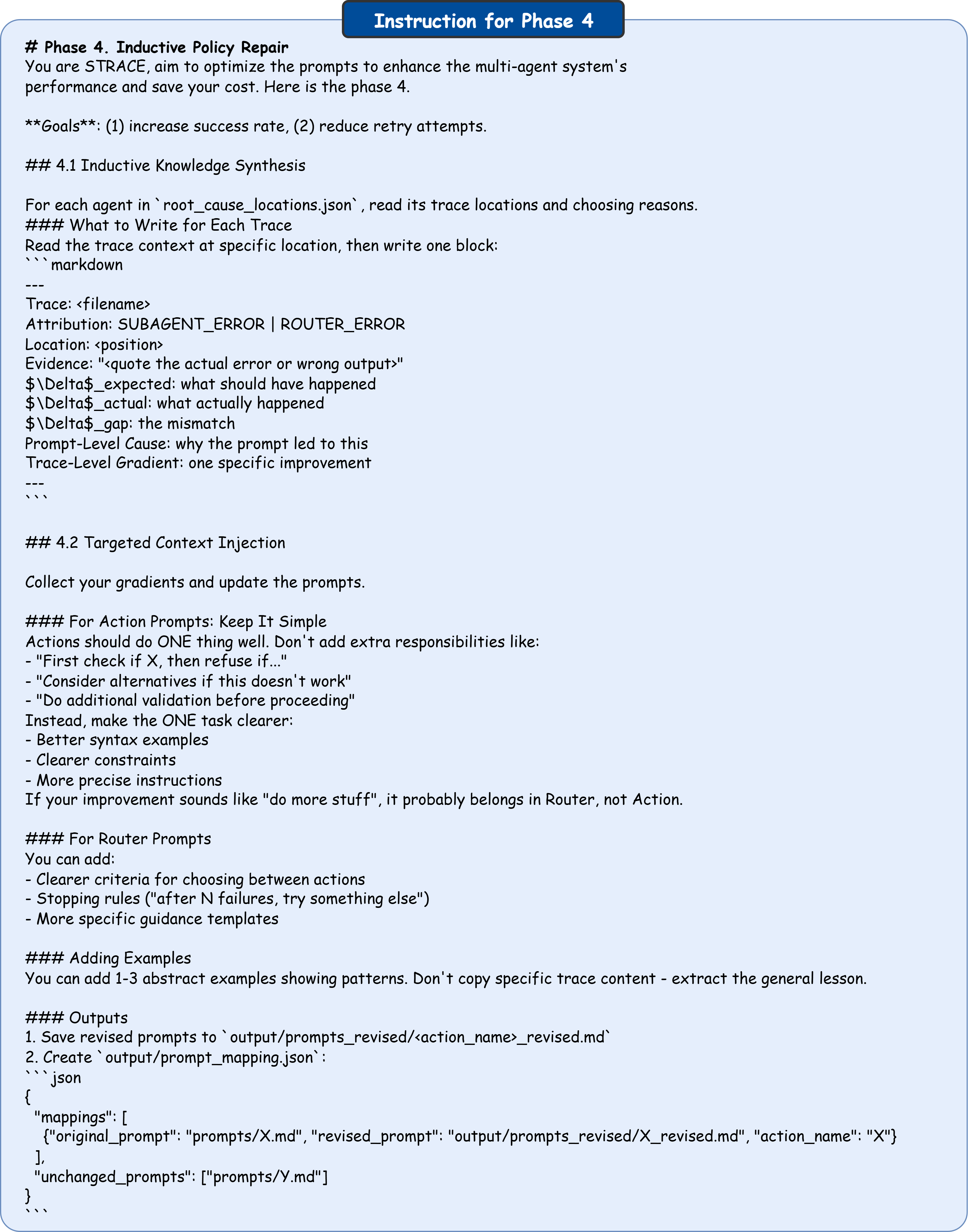}
  \caption{Instruction for Phase 4 -- Inductive Policy Optimization.}
  \label{fig:phase4_instruction}
  
\end{figure*}

For readability, we present simplified versions of the instructions for each phase in this appendix, as shown in Figures~\ref{fig:phase1_instruction}--\ref{fig:phase4_instruction}. To support reproducibility and reuse, the GitHub repository includes the full instruction templates together with a skill-based version that packages the same workflow for direct plug-in use.

\section{More Experiments Results}
\subsection{Stage-wise Cost Breakdown for the Ablation Study}
\label{app:cost_analysis}





Table~\ref{tab:ablation_cost} reports the stage-wise cost of STRACE in USD. Cost is estimated by tracking the input and output tokens consumed by optimizer-side LLM calls and multiplying them by the Claude Sonnet 4.5 API prices. For stage-wise cost, we apply the same calculation only to LLM calls within each phase, and the overall cost is the sum across phases. Phase~1 corresponds to \textit{Structural Modeling}, Phase~2 to \textit{Failure Pattern Mining and Trace Filtering}, Phase~3 to \textit{Causal Localization}, and Phase~4 to \textit{Inductive Policy Optimization}. A dash indicates the corresponding stage is removed in that variant. 


\paragraph{STRACE.}

The overall cost is 2.9563, dominated by Phase~4 (1.8052; 61.1\%), followed by Phase~3 (0.8515; 28.8\%). Phase~1 (0.1135; 3.8\%) and Phase~2 (0.1861; 6.3\%) contribute a relatively small fraction, indicating that the main cost driver is the downstream optimization stage, while diagnosis and slicing introduce only modest overhead.

\paragraph{w/o Structural Modeling (removing Phase~1).}
This variant removes Phase~1 to analyze the cost and performance impact of the dependency prior. Although Phase~1 itself is inexpensive in STRACE (0.1135, only 3.8\% of the total cost), removing it produces the largest performance degradation in the ablation study, reducing success rate from 56\% to 48\%. At the same time, the overall cost increases from 2.9563 to 5.1045, an increase of about 2.15. This indicates that without the dependency prior, the optimizer must spend substantially more effort inferring relationships among different components from traces alone, inspecting more trajectory context to reason about possible causal paths. Moreover, this inference is less reliable and can lead to incorrect failure paths. Thus, Phase~1 has low direct cost but plays an important role in making subsequent localization and repair both more accurate and more efficient.

\paragraph{w/o Trace Filtering (removing Phase~2).}
This variant removes Phase~2 and yields an overall cost of 8.4489, where Phase~4 alone increases to 7.602 (90.0\% of the total). In comparison, Phase~1 (0.1667; 2.0\%) and Phase~3 (0.6802; 8.1\%) remain small. The cost inflation is therefore primarily concentrated in Phase~4, consistent with the intuition that without diagnosis-driven localization, the subsequent optimization must operate with less targeted signals and incurs substantially higher downstream search/optimization overhead.

\paragraph{w/o Slicing (Current Node) (removing Phase~3).}
With slicing removed, the overall cost becomes 2.8762. The cost is still dominated by Phase~4 (2.0486; 71.2\%), while Phase~1 and Phase~2 increase to 0.2956 (10.3\%) and 0.5320 (18.5\%), respectively. This suggests that when compact dependency-preserving slices are unavailable, more computation shifts to earlier stages (especially diagnosis) and the optimization stage remains the major cost component.

\paragraph{w/o Slicing (Full Trace) (removing Phase~3).}
This variant also removes Phase~3 but uses the complete trajectory as context, resulting in an overall cost of 5.9276. Phase~4 rises sharply to 5.3179 (89.7\%), while Phase~1 and Phase~2 are 0.2500 (4.2\%) and 0.3597 (6.1\%). Compared to STRACE, the increase is almost entirely attributable to Phase~4, consistent with token/turn blow-up when long-horizon, unfiltered traces are directly used in downstream prompt optimization.

\begin{table*}[h]
  \caption{Average Turns of \sysname{} and VeruSAGE.}
  \label{tab:avg_turn}
  \begin{center}
    \begin{small}
        \resizebox{0.99\linewidth}{!}{%
        \begin{tabular}{c|ccccc}
        \toprule
          Method& IronKV&Memory Allocator&Node Replication&NRKernel&Storage\\
          \midrule
          VeruSAGE (Base Agent)&12.04&9.17&7.80&14.83&13.00\\
 VeruSAGE (\sysname{}-enhanced)& 8.42& 7.39& 4.00& 8.20&8.31\\
 \bottomrule
 \end{tabular}%
        }
    \end{small}
  \end{center}
\end{table*}

\begin{table*}[h]
  \caption{Performance on VeruSAGE-Bench of \sysname{} and other hands-off models.}
  \label{tab:hands_off_performance}
  \begin{center}
    \begin{small}
        \resizebox{0.99\linewidth}{!}{%
        \begin{tabular}{c|cccccc}
        \toprule
          SR& IronKV&Memory Allocator&Node Replication&NRKernel&Storage &Overall\\
          \midrule
 GPT-5& 75.0\%& 55.6\%& 90.0\%& 31.7\%&23.1\%&50.0\%\\
  \textbf{o4-mini (\sysname{}-enhanced)}&62.5\%&\textbf{88.9}\%&\textbf{100.0\%}&31.7\%&\textbf{61.5\%} &58.5\%\\
 Claude Sonnet 4& \textbf{79.2\%}& 66.7\%& 70.0\%& \textbf{46.3\%}& 46.2\%&\textbf{59.4\%}\\
 \bottomrule
 \end{tabular}%
        }
    \end{small}
  \end{center}
\end{table*}

\paragraph{Summary.}
Across variants, the largest cost variations are concentrated in Phase~4. Phase~2 (trace filtering) primarily reduces downstream optimization overhead by localizing failures, while Phase~3 (causal localization) avoids shifting costs to Phase~4 by providing compact, dependency-preserving context in place of full-trace inputs.

\subsection{Average Turns of VeruSAGE (\sysname{}-enhanced) Compared to the Base Agent}


In VeruSAGE, a \emph{turn} is the atomic unit of a repair attempt in the outer interaction loop, reflecting how many iterations are needed before the system succeeds or terminates. For most failure types, invoking a single sub-agent to propose and apply a fix counts as one turn. For assertion failures, VeruSAGE first triggers an \texttt{assertion\_reasoning\_pipeline} sub-agent that analyzes the error and selects a specialized downstream sub-agent; in this case, the \texttt{assertion\_reasoning\_pipeline} together with its selected sub-agent are treated as one turn, since they form a coupled decision--execution step within the same outer-loop attempt.

\textbf{Average Turns} complements Success Rate by characterizing efficiency among successful (and attempted) repairs. Fewer turns indicate faster convergence with fewer outer-loop iterations, which typically corresponds to fewer LLM calls, lower latency, and reduced interaction overhead under the same stopping criteria (i.e., stopping early upon success or terminating at the maximum allowed attempts). For each task instance, the number of turns is counted until success. If the agent fails to solve within the allowed attempts, the maximum-turn cutoff is recorded. The reported value is then averaged across the evaluation set.


As shown in Table~\ref{tab:avg_turn}, VeruSAGE (\sysname{}-enhanced) consistently requires fewer turns than the VeruSAGE base agent across all five VeruSAGE-Bench categories, indicating faster convergence in the outer repair loop. The reduction is most pronounced on the more challenging benchmarks (e.g., IronKV and NRKernel), while the remaining categories also exhibit steady decreases. Overall, these results suggest that \sysname{}-enhanced prompting improves per-turn decision quality and action effectiveness within VeruSAGE, thereby reaching a valid repair with fewer interaction rounds and lower iterative overhead under the same stopping criteria.

\subsection{\sysname{} Performance on VeruSAGE-Bench Compared with Hands-off Agents}


We additionally compare \sysname{} with hands-off agents reported in VeruSAGE. In VeruSAGE, the hands-off approach instantiates a generic coding agent (a CLI agent) with a lightweight prompt that primarily enforces anti-cheating constraints, while granting the model direct tool access to (i) running Verus, (ii) running a cheat checker, and (iii) inspecting the Verus standard library (\texttt{vstd}) as needed. This setting intentionally provides little verification-specific tutoring, and relies on the underlying coding capability of frontier models together with tool feedback from the verifier. \cite{verusage}

We report two hands-off baselines based on frontier models: GPT-5 and Claude Sonnet 4. Their hands-off numbers are collected from the VeruSAGE study and used as reference results under the same VeruSAGE-Bench suite and evaluation protocol. \sysname{} is evaluated on the same benchmark split and compared against these hands-off agents accordingly.


As shown in Table~\ref{tab:hands_off_performance}, \sysname{} optimized VeruSAGE achieves an overall success rate that is competitive with strong hands-off agents. In particular, \sysname{} outperforms the hands-off GPT-5 baseline in overall SR, and reaches a comparable overall level to Claude Sonnet 4 (58.5\% vs 59.4\%), despite using a substantially smaller model o4-mini. The gains are most evident on Memory Allocator, Node Replication, and Storage, indicating that our learned, task-specific guidance can substantially improve repair reliability in several benchmark categories. Meanwhile, the gap on IronKV and NRKernel suggests that some failure modes still benefit from stronger general reasoning and long-horizon robustness. These results demonstrate that \sysname{} can meaningfully close the performance gap between lightweight models and frontier hands-off agents on VeruSAGE-Bench.

\section{Evolution Examples in VeruSAGE Task}
\label{app:generalized_rules_details}

In this section, we present the actual generalized rules synthesized from the VeruSAGE benchmark for the \texttt{assertion\_reasoning\_pipeline} module. 
Crucially, \sysname{} de-contextualizes specific trace observations (e.g., \textit{"Sub-agent A failed on input X"}) into transferable heuristics reusable across varying proof states. In particular, the revised prompt distills three types of reusable rules:

\begin{enumerate}[itemsep=0pt, parsep=0pt, topsep=2pt]
    \item \textbf{Failure-aware stopping condition and action switching.}
    \begin{itemize}[itemsep=0pt, parsep=0pt, topsep=2pt]
        \item \textbf{Track repeated failures:} Maintain awareness of how many times each action type has been attempted and rejected under the current proof context.

        \item \textbf{Stopping condition:} If an action has been attempted 2--3 times and is consistently rejected (especially due to syntax errors or does not fix the target assertion), \textbf{do not} select that action again in the same context.

        \item \textbf{Action switching:} Instead, switch to a different secondary action from the candidate set; if all reasonable options are exhausted, explicitly mark the issue as requiring human intervention.

        \item \textbf{Example:} If INDUCTION fails three times with syntax errors, switch to alternatives such as USELEMMA or CASE\_ANALYSIS rather than retrying INDUCTION again.
    \end{itemize}

    \item \textbf{Sub-agent-selection boundaries and actionable guidance.}
    \begin{itemize}[itemsep=0pt, parsep=0pt, topsep=2pt]
        \item \textbf{General requirement:} The guidance field must be specific and actionable (avoid generic suggestions); include concrete anchors from the visible code and the current goal structure.

        \item \textbf{USELEMMA:} Identify specific helper lemmas/proof functions available in the file and state what each lemma establishes. Do \textbf{not} select USELEMMA if no relevant lemmas exist; prefer actions that construct the proof directly (e.g., CASE\_ANALYSIS, INDUCTION, COMPUTE).

        \item \textbf{CASE\_ANALYSIS:} Specify the exact variable or expression to split on and what each branch should establish (e.g., which inequality or domain condition is proven in each case).

        \item \textbf{INDUCTION:} State the inductive structure (base case and inductive hypothesis) and how the hypothesis connects to the current goal.

        \item \textbf{EXTENSIONAL\_EQUALITY:} Indicate which collection equality/property requires element-wise reasoning and what domain/length conditions need to be established.

        \item \textbf{SEQSETMAP:} List the 2--3 most relevant Seq/Set/Map lemmas to invoke; avoid invoking an overly broad lemma set to prevent timeouts.
    \end{itemize}

    \item \textbf{Multi-round planning strategy.}
    \begin{itemize}[itemsep=0pt, parsep=0pt, topsep=2pt]
        \item \textbf{Build on progress:} Explicitly reference what has been established by previously accepted fixes and explain how the next action extends that progress.

        \item \textbf{Develop a short proof plan:} For complex goals, outline a minimal multi-step plan (e.g., establish a key auxiliary fact, prove the main implication, connect to the final goal).

        \item \textbf{Avoid random exploration:} Do not switch between unrelated actions without a clear logical progression; each action should advance the proof toward the current failing goal.
    \end{itemize}
\end{enumerate}

\end{document}